\documentclass[journal]{IEEEtran}
\ifCLASSINFOpdf
\else
\fi
\usepackage{graphics} 
\usepackage{amsmath} 
\usepackage{amssymb}  
\usepackage{amsfonts}
\usepackage{xcolor}
\usepackage{epsfig}
\usepackage{graphicx}
\usepackage{epstopdf} 
\usepackage[linesnumbered,ruled,vlined]{algorithm2e}
\SetNlSty{bfseries}{\color{black}}{}
\usepackage{bm}
\usepackage{gensymb}
\usepackage{verbatim}
\usepackage{cite}

\usepackage{paralist}
\usepackage{flushend}
\usepackage{booktabs}
\usepackage{multirow}
\usepackage{stfloats}
\usepackage{caption}
\usepackage{subcaption}
\usepackage{mathtools}

\usepackage{array}

\begin{document}
%
\title{Conditional Generative Adversarial Networks for Optimal Path Planning}

\author{Nachuan Ma $^{1\dagger}$, Jiankun Wang $^{1\dagger}$, and Max Q.-H. Meng$^{2}$, \emph{Fellow, IEEE}
	\thanks{$^{\dagger}$Equal contributions.}
	\thanks{$^{1}$Nachuan Ma and Jiankun Wang are with the Department of Electronic and Electrical Engineering of the Southern University of Science and Technology in Shenzhen, China,
		{\tt\small \{manc@mail.,wangjk@\}sustech.edu.cn}}%
	\thanks{$^{2}$Max Q.-H. Meng is with the Department of Electronic and Electrical Engineering of the Southern University of Science and Technology in Shenzhen, China, on leave from the Department of Electronic Engineering, The Chinese University of Hong Kong, Hong Kong, and also with the Shenzhen Research Institute of the Chinese University of Hong Kong in Shenzhen, China. (\emph{Corresponding author: {\tt\small max.meng@ieee.org}})}%
}

%



\maketitle

\begin{abstract}
Path planning plays an important role in autonomous robot systems. 
Effective understanding of the surrounding environment and efficient generation of optimal collision-free path are both critical parts for solving path planning problem. 
Although conventional sampling-based algorithms, such as the rapidly-exploring random tree (RRT) and its improved optimal version (RRT*), have been widely used in path planning problems because of their ability to find a feasible path in even complex environments, they fail to find an optimal path efficiently. 
To solve this problem and satisfy the two aforementioned requirements, we propose a novel learning-based path planning algorithm which consists of a novel generative model based on the conditional generative adversarial networks (CGAN) and a modified RRT* algorithm (denoted by CGAN-RRT*). 
Given the map information, our CGAN model can generate an efficient possibility distribution of feasible paths, which can be utilized by the CGAN-RRT* algorithm to find the optimal path with a non-uniform sampling strategy. 
The CGAN model is trained by learning from ground truth maps, each of which is generated by putting all the results of executing RRT algorithm 50 times on one raw map. 
We demonstrate the efficient performance of this CGAN model by testing it on two groups of maps and comparing CGAN-RRT* algorithm with conventional RRT* algorithm.
\end{abstract}

	


\begin{figure}[!htb]
	\captionsetup{font={small}}
	\setlength{\abovecaptionskip}{0.cm}
	\begin{subfigure}[b]{.4\linewidth}
		\includegraphics[width=1.05\linewidth]{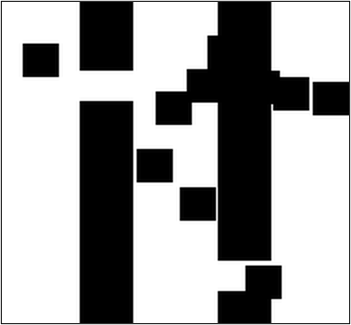}
		\caption{ }
	\end{subfigure}
	\hfill
	\begin{subfigure}[b]{.4\linewidth}
		\includegraphics[width=1.05\linewidth]{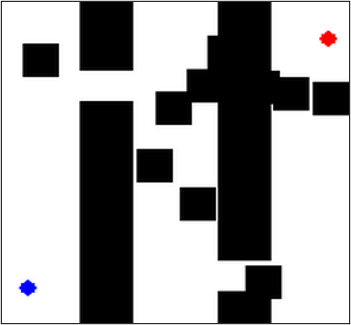}
		\caption{ }
	\end{subfigure}
	\hfill
	\begin{subfigure}[b]{.4\linewidth}
		\includegraphics[width=1.05\linewidth]{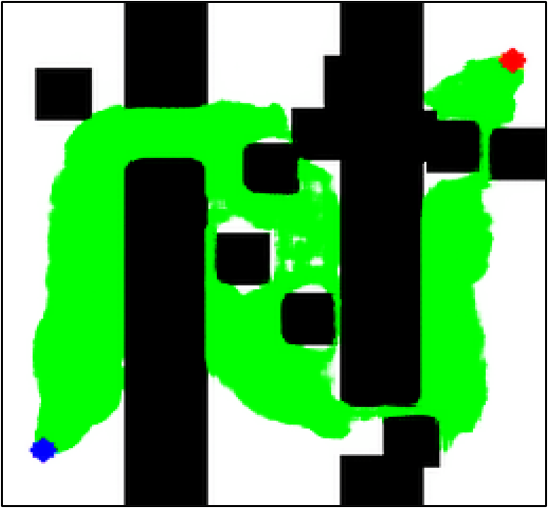}
		\caption{ }
	\end{subfigure}
	\hfill
	\begin{subfigure}[b]{.4\linewidth}
		\includegraphics[width=1.05\linewidth]{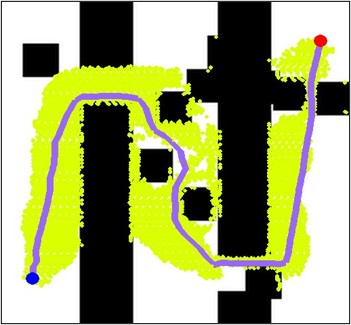}
		\caption{ }
	\end{subfigure}
	\caption{ Four stages of the proposed CGAN-RRT* algorithm. 
		(a): The original map. 
		(b): The map(a) added randomly assigned start point and endpoint. 
		(c): The map(b) added predicted possibility distribution of feasible paths generated from the CGAN model.
		(d): The map(c) added the optimal path generated from the CGAN-RRT* algorithm.}
	\label{fig1:two_grid_worlds}
\end{figure}

%
\IEEEpeerreviewmaketitle

\section{Introduction}
\label{sec:introduction}
\IEEEPARstart{P}{ath} planning is an essential component for autonomous mobile robots. 
Its performance directly determines the success rate of robot tasks and robustness of robot systems. 
The objective of the path planning problem is to generate a collision-free path for robots from a start state to a goal state while avoiding a number of moving or static obstacles. 
Over the past few decades,  researchers have proposed a lot of path planning algorithms, 
such as Artificial Potential Field Method (APF) \cite{khatib1986real}, 
Probabilistic Roadmap Method (PRM) \cite{kavraki1996probabilistic}, 
Rapidly-Exploring Random Tree (RRT) \cite{lavalle2001randomized}, 
A* algorithm \cite{hart1968formal} and so on. 
By utilizing a virtual force method, 
the APF transforms the movement of robots in the environmental space into the virtual potential field to guide the movement of robots. 
However, it suffers from easily falling into a local minimum.  
The A* algorithm adopts a heuristic function and computes its value at each node within configuration space to obtain the optimal path.
But it tends to consume much time and huge memory usage with the increase of the size of configuration space. 
The RRT algorithm iteratively constructs trees to connect samples drawn from the given sampling distributions in a short time.
Then a feasible path can be found by traversing the tree from the start node to goal node.
The PRM algorithm randomly samples points to establish a graph which can be solved by a typical graph search algorithm.
It utilizes a local planner to connect random samples from free space.

In the aforementioned path planning algorithms, 
sampling-based algorithms such as the RRT and PRM algorithms are widely applied for autonomous mobile robots because of their probabilistic completeness and good scalability.
Besides, sampling-based path planning algorithms do not require a elaborate modeling of the environment. 
However, both RRT and PRM algorithms cannot guarantee the optimal path.
To make the initial path converge to an optimal path, the RRT* algorithm \cite{karaman2011sampling} is proposed.
It can be viewed as a significant improvement over the RRT algorithm.
But the RRT* algorithm converges slowly to the optimal path because planners probabilistically draw random samples from a uniform distribution.
It is also constrained by the quality of the initial path.
To the best of our knowledge, the existing RRT* algorithms cannot solve a path planning problem quickly especially when encountering challenging environments such as a environment containing narrow passages or many turns. 

To overcome the limitations of the RRT* algorithm, researchers have proposed some learning-based path planning algorithms which change the sampling strategy of the RRT* algorithm \cite{gammell2014informed} \cite{wang2020neural} \cite{akgun2011sampling}.
Inspired by these improved sampling-based algorithms,  
this article aims to construct a neural network model that can quickly predict the probability distribution of feasible paths on a map.
And the predicted possibility distribution is utilized to guide the sampling process of the RRT* algorithm. 
The problem can be viewed as ``translating" an input image into a corresponding output image. 
In the field of computer vision, researchers have already achieved significant progress in the problem of predicting pixels to pixels with convolutional neural networks (CNNs) \cite{hertzmann2001image}. 
CNNs learn to minimize a loss function of Euclidean distance between predicted and ground truth pixels. 
However, it will tend to generate blurry results because Euclidean distance is minimized by averaging all plausible outputs. 
To make the output indistinguishable from the reality, the Generative Adversarial Networks (GANs) \cite{goodfellow2014generative} can be applied in this problem. 
GANs train a generator to generate outputs which cannot be distinguished from the real images by a synchronously training discriminator which tries to detect fake maps. 
For image-to-image prediction tasks, GANs can be modified in the conditional setting. 
The conditional GANs (CGANs) condition on an input image and generate a corresponding output image.
CGANs are more suitable for the path planning problem than GANs because they can direct the data generation process under certain conditions. CAGNS can be utilized to generate predicted possibility distribution of feasible paths on given maps.

In this article,  we propose a novel learning-based path planning algorithm which trains a CGAN model to obtain the predicted possibility distribution of feasible paths. 
The CAGN model is trained by learning from $12,000$ pairs of input maps and corresponding ground truth maps.
The input maps contain randomly start and goal points, 
while the ground truth maps contain $50$ feasible paths generated from the RRT algorithm.
For a given path planning problem, the proposed CGAN model can quickly predict the probability distribution of feasible paths on the map. 
Then, the predicted possibility distribution is utilized to guide the sampling process of the RRT* algorithm. 
The results of the model are promising and show that it can improve the performance of the RRT* algorithm significantly. 
This new model can also be applied to other sampling-based algorithms in future research.

To summarize, the key contributions of the article include:
\begin{IEEEitemize}
  \item A conditional-GAN model to predict the probability distribution of the feasible paths for different types of maps;
  \item A novel optimal path planning algorithm which combines a CGAN model and the RRT* algorithm, denoting as the CGAN-RRT*;
  \item Case studies to demonstrate the effectiveness and efficiency of the proposed CGAN-RRT* algorithm.
\end{IEEEitemize}

The remainder of this paper is organized as follows. 
We introduce some related work concerning the GANs model and RRT algorithms in Section~\ref{sec:Related work}.
In Section~\ref{sec:Preliminaries}, we formulate the path planning problem and explain the RRT and RRT* algorithms.
Then Section~\ref{sec:DCGAN-RRT* Algorithm} presents the details of the proposed CGAN-RRT* algorithm.
A number of simulation experiments are conducted in Section~\ref{sec:Simulation Results} to compare the performance of CGAN-RRT* with that of RRT*.
Finally, Section~\ref{sec:conclusion} concludes this article and discusses future work.

\section{Related work}
\label{sec:Related work}
\subsection{Improved Algorithms based on the RRT algorithm}
For decades, path planning has been a very active topic in robotics research. 
There are many well-established algorithms. 
As some general path planning techniques have been discussed in Section \ref{sec:introduction}, we will focus on improved algorithms based on the RRT algorithm.

In \cite{rickert2014balancing}, The exploring/exploiting tree (EET) is proposed by balancing exploitation and exploration during path planning.
Although it improves the computational efficiency of conventional sampling-based algorithms, it lacks probabilistic completeness.  
Yershove \textsl{et al.} \cite{yershova2005dynamic} propose the dynamic-domain RRT algorithm which improves the performance of the RRT algorithm in several motion planning problems. 
It adopts a new sampling strategy which limits the expansion of nodes near obstacles during the sampling process. 
However, the dynamic-domain RRT algorithm is not convenient as a new parameter of it needs to be tuned carefully.
To automatically tune the new parameter, a extension of the dynamic-domain RRT algorithm is proposed in \cite{jaillet2005adaptive}.
It improves the performance of the dynamic-domain RRT algorithm by adapting the region of influence of each node. 
By utilizing the initial path generated from the A* algorithm to guide the sampling process of the RRT* planner, 
Brunner \textsl{et al.} \cite{brunner2013hierarchical} propose the A*-RRT* algorithm. 
It improves the convergence speed of the RRT* algorithm.
However, both algorithms in \cite{jaillet2005adaptive} and \cite{brunner2013hierarchical} suffer from consuming too much time when the size of configuration space increases. 
 

Recently, deep learning and reinforcement learning methods have been applied in robotic path planning problems. 
Reinforcement learning method has achieved good performance in decision-making problems and deep learning method is suitable for image recognition.
By modifying the coefficient of the Bellman equation, Zhang \textsl{et al.} \cite{zhang2012reinforcement} propose an improved method to accelerate the convergence of the Q-learning algorithm in path planning tasks.
Some scholars obtain good path planning results by combining deep learning methods and reinforcement learning methods.
Dubey \textsl{et al.} \cite{dubey2013path} utilizes images generated by the Q-learning algorithm as the training data. 
Then the training data is input to a deep-earning neural network to learn a path.
However, both algorithms in \cite{zhang2012reinforcement} and \cite{dubey2013path} are constrained by low convergence speed and much time cost of the Q-learning algorithm.
In \cite{ichter2018learning}, the conditional variational auto-encoder (CVAE) algorithm for robot motion planning is proposed. 
It utilizes a nonuniform sampling strategy which learns from demonstrations of successful motion plans to guide the sampling process. 	
By implementing a CNN model, Wang \textsl{et al.} \cite{wang2020neural} propose the Neural-RRT* algorithm. 
It utilizes the predicted possibility distribution generated from the CNN model to guide the sampling process of the RRT* algorithm.
Though the Neural-RRT* algorithm achieves better performance than the informed RRT* and RRT* algorithms,
the predicted sampling regions generated from the CNN model may be discontinuous sometimes.

\subsection{Conditional GANs}
In recent years, applying Generative Adversarial Network (GAN) in the conditional setting has gained a lot of attention in computer vision field. 
Mirza  \textsl{et al.} \cite{mirza2014conditional} introduce a condition version of GAN. 
The model can generate MNIST digests conditioned on class labels. 
In \cite{gauthier2014conditional}, Gauthier further applies the Conditional Generative Adversarial Network (CGAN) in face recognition.
Positive results have been obtained in utilizing the combined conditional data to control particular face attributes from the model. 
Denton \textsl{et al.} \cite{denton2015deep} present a Laplacian Pyramid of Adversarial Networks to produce high-quality samples of natural images.
The model utilizes a cascade of convolutional networks within a Laplacian pyramid framework. 
In \cite{wang2016generative}, the Style and Structure Adversarial network (S²-GAN) is presented. 
The model utilizes the Structure-GAN to generate a surface normal map. 
Then the Style-GAN translates the normal map into the corresponding 2D image. 
Mathieu \textsl{et al.} \cite{mathieu2015deep} evaluates different loss functions and illustrates that generative adversarial training can be effective for future frame prediction. 
For product photo generation problem, Yoo \textsl{et al.} \cite{yoo2016pixel} propose a pixel-level domain converter. 
It translates the information in the source domain into a pixel-level image while preserving the semantic meaning.

To the best of our knowledge, there exist few applications for the path planning problem by utilizing GANs.
Different from the aforementioned path planning algorithms, in this article, we propose a CAGN model to quickly predict the probabilistic distribution of feasible paths on the map. 
Our method also differs from the aforementioned CGAN model works in the architectural choice. 
A ``U-net" - based architecture is utilized in our generator \cite{ronneberger2015u}, which is a popular version of GAN in image-to-image tasks.
We show that this approach is effective in the path planning problem by experimental simulations in Section V.

\section{Preliminaries}
\label{sec:Preliminaries}
This section starts with the formulation of the path planning problem and an explanation of the RRT and RRT* algorithms is presented in Section \ref{sec:Preliminaries}-B. 
The path generation results of both algorithms are shown in Fig. \ref{fig2}.

\subsection{Path Planning Problem and Related Terminologies}
In this section, the basic path planning problem is formalized. 
We denote $\mathcal{X} \in$ $\mathbb{R}^{d}$ as the state space. 
Let $\mathcal{X}_{free}$ be the free space, and $\mathcal{X}_{obs} = \mathcal{X} / \mathcal{X}_{free}$ is denoted as the obstacle space. 
We denote $x_{init}$ and $x_{goal}$ as the initial state and the goal state, respectively. 
They both belong to $\mathcal{X}_{free}$. 
In addition, $\mathcal{X}_{goal}$ $\in$ $\mathcal{X}_{free}$ denotes the goal region. 
Then, the path planning problem can be expressed as to generate a feasible path $\delta(t) \in \mathcal{X}_{free}$ for $t \in [0,t_{g}] $ that satisfies the robot dynamics constraints.
The path starts at the initial state $\delta(0) = x_{init}$ and ends at the goal state $\delta(t_{g})$ $\in$ $\mathcal{X}_{goal}$, where $\mathcal{X}_{goal} = \{{x \in \mathcal{X}_{free}}, \big| ||x- x_{goal}|| <r\}$, and $r$ is a positive real number.

We also employ the definition of \emph{Optimal Path} throughout the rest of this article:

$\emph{Optimal\;Path}$: Given a path planning problem, the cost function $L$ is designed for obtaining path length. 
If we find a feasible path that can minimize the cost function $L(\delta(t))$, where $\delta(t) \in \mathcal{X}_{free}$, we define it as the optimal path $\delta(t)^*$.

The definitions which are utilized in the RRT algorithm are shown as follows:

$\emph{Samplefree}$: Sample a random state from the state space $\mathcal{X} \in$ $\mathbb{R}^{d}$.

$V$: The set of vertices in the sampling tree.

$E$: The set of edges between the vertices in the sampling tree.

$\emph{Nearest}(V,x)$: Find the nearest node from $V$ to the point $x$ by utilizing a Euclidean distance.

$\emph{Steer}(x_{1},x_{2})$: Steer from $x_{1}$ to $x_{2}$ along the path $\delta(t)$.

$\emph{ObstacleFree}(\delta(t))$: Determine whether the path $\delta(t)$ is feasible and collision-free.

\subsection{The RRT and RRT* algorithms}
\begin{algorithm}[h]
	\normalsize
	\label{RRT}
	\caption{The RRT Algorithm}
	$V$ $\gets$ \{$x_{init}$\}; $E$ $\gets$ $\emptyset$;
	\\
	\For {$i$=1,. . .,$n$}{{$x_{rand}$ $\gets$ \text{SampleFree}$_{i}$};
		\\	
		{$x_{nearest}$} $\gets$ \text{Nearest}($G=(V,E)$,{$x_{rand}$});
		\\
		{$x_{new}$} $\gets$ \text{Steer}({$x_{nearest}$},{$x_{rand}$});
		\\
		\If{\text{ObstacleFree}$({x_{nearest}},{x_{new}})$}{$V$ $\gets$ $V$ $\cup$ \{{$x_{new}$}\}; $E$ $\gets$ $E$ $\cup$ \{{($x_{nearest}$,$x_{new}$)}\} 
		}
	}
	\textbf{Return} ($G=(V,E)$)
\end{algorithm}
The pseudo-code for the RRT algorithm is presented in Algorithm $1$. 
It is mainly designed for single-query applications and appropriate for dealing with high dimension problems. 
The RRT algorithm randomly expands nodes in the collision-free space and incrementally builds a tree of feasible paths rooted at the initial state. 
It begins with an empty edge set $E$ and a vertex set $V$ including the initial state $x_{init} \in \mathcal{X}_{free}$.
During the execution of the algorithm, a node $x_{rand} \in \mathcal{X}_{free} $ is sampled randomly at every iteration.
Then the nearest node $x_{nearest}$ of the existing vertex set $V$ is connected to the new sampling node $x_{rand}$.
If the connection works, the edge set $E$ adds $(x_{nearest}, x_{rand})$ and the vertex set $V$ adds the new sampling node $x_{rand}$.
Finally, the iteration is stopped when the expanding tree contains a new sampling node in the goal region and the algorithm returns the edge set $E$ and the vertex set $V$.

\begin{figure}[!htb]
	\setlength{\abovecaptionskip}{0.cm}
	\begin{subfigure}[b]{.4\linewidth}
		\includegraphics[width=1\linewidth]{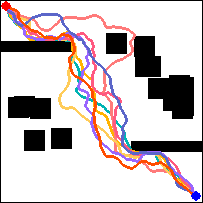}
		\caption{ }
		\label{fig2:a}
	\end{subfigure}
	\hfill
	\begin{subfigure}[b]{.4\linewidth}
		\includegraphics[width=1\linewidth]{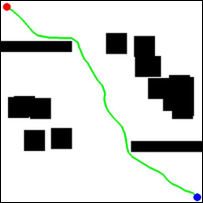}
		\caption{ }
		\label{fig2:b}
	\end{subfigure}
	\caption{Comparison between the RRT and RRT* algorithms.
		(a): $8$ paths generated by the RRT algorithm within one map.
		(b): The optimal path generated by the RRT* algorithm.}
	\label{fig2}
\end{figure}

The RRT algorithm is popular for its probability completeness, good scalability, and high efficiency. 
It can guarantee the feasibility of a solution if the map exists a path between the initial state and the goal region. 
However, though the RRT algorithm can find a feasible path in a short time, it cannot obtain the optimal path. 
Actually, on the same map, the RRT algorithm tends to generate different paths between the start and goal points as it samples vertices randomly. 
One example is shown in Fig. \ref{fig2}. 
The first image shows $8$ paths generated from the RRT algorithm in the same environment. 
We can find that some paths have better quality with fewer nodes and shorter length than other paths. 
It means that the RRT algorithm performs more efficiently when generating these paths. 
However, all 8 paths are generated randomly from the start point to the goal point. As a result, the RRT algorithm cannot guarantee the optimal solution and lacks stability.

To generate the optimal path, the RRT* algorithm is proposed. 
It is a variant of the RRT algorithm and also inherits the merit of the A* algorithm. 
The main difference from the RRT algorithm is that the RRT* algorithm adds a procedure of $\mathit{Rewire}$. 
It is an important procedure for the RRT* algorithm to determine which neighbor vertex has the shortest path length through $x_{new}$. 
From Fig. \ref{fig2} (b), we can find that the path obtained by the RRT* algorithm has the best performance with shorter path length. 
Although obtaining the optimal path is essential in some applications, the RRT* algorithm usually consumes much time and huge memory usage to converge to the optimal path. 
In order to accelerate the convergence speed of the RRT* algorithm, we propose the DCGAN-RRT* algorithm in Section~\ref{sec:DCGAN-RRT* Algorithm}. 

\section{CGAN-RRT* Algorithm}
\label{sec:DCGAN-RRT* Algorithm}
In this section, we first introduce the objective of our conditional GAN model in Section \ref{sec:DCGAN-RRT* Algorithm}-A.
Then the structure and loss function of the generator and discriminator is presented in Section \ref{sec:DCGAN-RRT* Algorithm}-B and \ref{sec:DCGAN-RRT* Algorithm}-C. 
Finally, we introduce the CGAN-RRT* algorithm in Sections IV-D.

\begin{figure*}[h]
	\centering
	\captionsetup{font={small}}
	\setlength{\abovecaptionskip}{0.cm}
	\begin{subfigure}[b]{.85\linewidth}
		\includegraphics[width=1.1\linewidth]{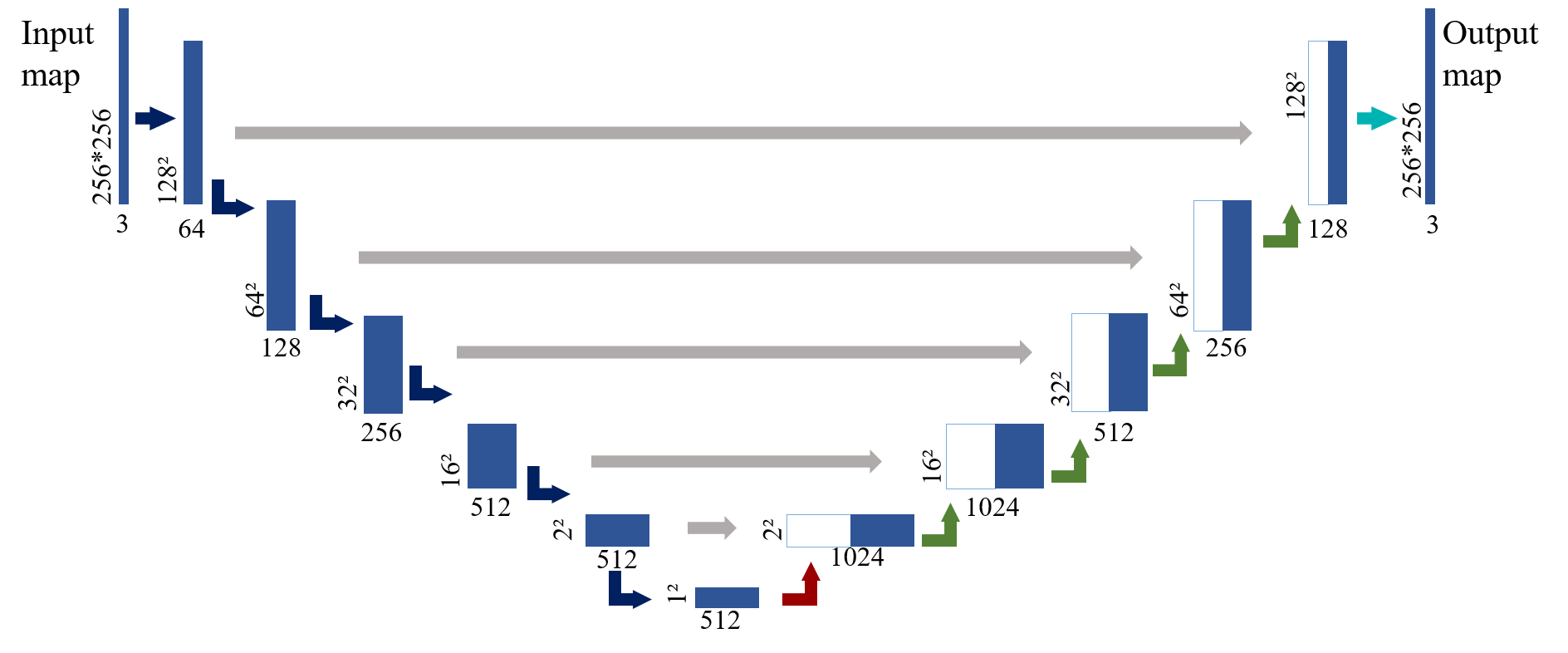}
		\caption{ }
	\end{subfigure}
	\hfill
	\begin{subfigure}[b]{.85\linewidth}
		\includegraphics[width=1.1\linewidth]{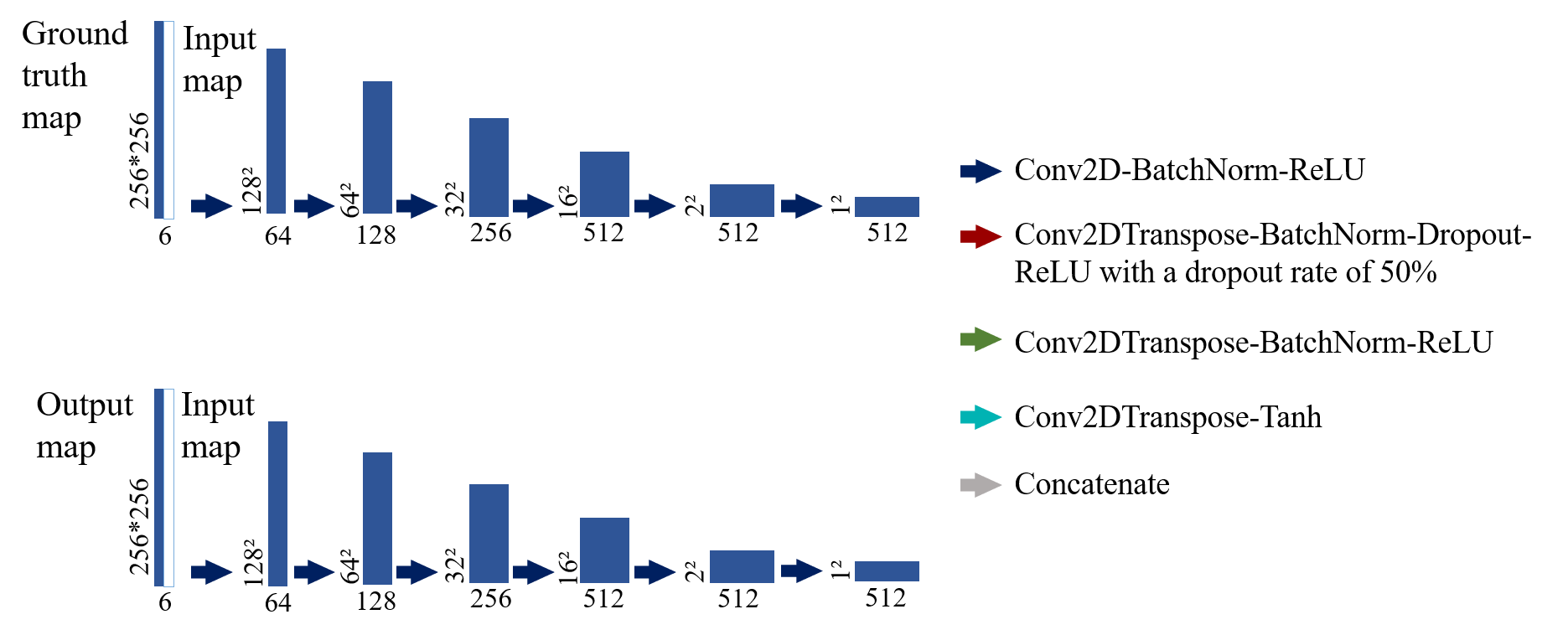}
		\caption{ }
	\end{subfigure}
	\caption{Illustration of the proposed CGAN model.
		Blue boxes represent multi-channel feature maps and white boxes represent copied feature maps.
		(a) The architecture of the generator with 'u-net' structure.
		(b) The architecture of two discriminators with two pairs of input.}
	\label{fig3}
\end{figure*}


\subsection{Objective}
Usually, the Generative Adversarial Networks (GANs) are designed to learn a mapping from random noise $z$ to the objective output image $y$.
Then we have $\{G:z \rightarrow y\}$, where $G$ represents the generator of a GAN.
It is trained to generate output images that can confuse the discriminator $D$. 
$D$ is trained to distinguish the $fake$ data produced by $G$ from the $real$ data.

The objective of a GAN can be expressed as:
\begin{equation}\label{key}
\begin{aligned}
\mathcal{L}_{GAN}(G,D) = & \mathbb{E}_{y}[logD(y)] +
\\ 
& \mathbb{E}_{z}[log(1-D(G(z))].
\end{aligned}
\end{equation}

The generator is trained to minimize the objective, while the discriminator is trained to maximize it. 
Then the problem of an unconditional GAN is represented as $\theta^{*} = {\arg \,\min_{G} \max_{D}}\mathcal{L}_{GAN}(G,D)$.

In contrast to the unconditional GAN, the conditional-GAN learns a mapping from observed image $o$ and random noise $z$ to output image $y$.
As the discriminator observes $o$, the objective function of the conditional-GAN is expressed as:
\begin{equation}\label{key}
\begin{aligned}
\mathcal{L}_{cGAN}(G,D) = & \mathbb{E}_{o,y}[logD(o,y)] +
\\ 
& \mathbb{E}_{o,z}[log(1-D(o,G(o,z))].
\end{aligned}
\end{equation}

We add $L1$ loss to encourage less blurring, which is the mean absolute error between the generated image and the target image.
Then the final objective of our conditional-GAN model is represented as $\theta^{*} = {\arg \,\min_{G} \max_{D}}[\mathcal{L}_{cGAN}(G,D) + \lambda\mathcal{L}_{L1}(G)]$,
where $\lambda$ is a weight coefficient. In this article, we set $\lambda=100$.

For the path generation problem, observed image $o$ represents the original map with start and goal points.
The $fake$ data represents maps with predicted possibility distribution of feasible paths, which are produced by the generator.
The $real$ data represents maps with $50$ feasible paths generated from the conventional RRT algorithm. 

\subsection{Generator Architectures}
The generator aims to translate a high-resolution input map to another high-resolution output map.
The input maps represent original maps with randomly start and goal points, 
while the output maps represent maps with predicted possibility distribution of feasible paths.
Both corresponding input and output maps share the same start point, goal point and the structure of the original map. 
They just differ in the surface where output maps have the predicted possibility distribution of feasible paths.
Our generator architecture is designed under this consideration.
The proposed model is shown in Fig. \ref{fig3}.
All convolution modules are $4 \times 4$ spatial filters applied with stride $2$.
Convolution modules down-sample by a factor of $2$ in the encoder and up-sample by a factor of $2$ in the decoder.
The dimension of both input and output maps is $(256, 256, 3)$, which means that they are pictures of $(256, 256)$ size with $3$ channels.

For the encoding stage, we use modules of the form Convolution-Batchnorm-ReLU.
The output of each encoding layer is called a feature map. 
Low convolutional layers extract low-level feature maps and they are fed into a higher convolution layer to extract a high-level feature map.
If we denote each encoding layer with k filters as $\mathit{C_{k}}$, the encoder architecture can be expressed as:
$\mathit{C_{64}}$-$\mathit{C_{128}}$-$\mathit{C_{256}}$-$\mathit{C_{512}}$-$\mathit{C_{512}}$-$\mathit{C_{512}}$-$\mathit{C_{512}}$-$\mathit{C_{512}}$.
The Batchnorm layer aims to perform the normalization operation for each training mini-batch.  
It can accelerate the neural network training process and deal with the problem of parameter initialization \cite{ioffe2015batch}.
The rectified linear unit (ReLU) activation layer is widely used in neural networks, which can help avoid the over-fitting problem \cite{nair2010rectified}.
It assigns zero output for negative input.
The activation function of it is defined as $f(u)=\max(0,u)$, where $u$ is the input value.
Leaky-ReLU is a variant of ReLU. 
It translates negative input into output in the range of $(0,1)$, that is $f(u)=\max(\eta u,u)$.
All ReLU activation layers in the encoder are leaky, with slope $\eta=0.2$.

For the decoding stage, we use modules of the form ConvTranspose-Batchnorm-dropout-ReLU.
Let $\mathit{T_{k}}$ denote the decoding layers with k filters and $\mathit{TD_{k}}$ denote the decoding layers with a dropout rate of $50\%$.
Then the decoder architecture can be expressed as:
$\mathit{TD_{512}}$-$\mathit{TD_{512}}$-$\mathit{TD_{512}}$-$\mathit{T_{512}}$-$\mathit{T_{256}}$-$\mathit{T_{128}}$-$\mathit{T_{64}}$.
For a number of conditional-GAN models, random noise $z$ is added into the input of the generator to avoid generating deterministic outputs.
However, the generator tends to ignore the random noise, which makes this strategy ineffective.
In our decoder architecture, a dropout layer is utilized to act as noise and ensure randomness, which randomly sets a half of input units to zero at each step \cite{hinton2012improving}. 
We apply the dropout layer in the first three decoding layers at both training and test time.
All ReLU layers in the decoding stage are not leaky.

We adapt ``U-NET'' as the backbone of our generator network. 
It is designed for shuttling abundant information between the input and output directly across the network.
In the generator architecture, we have $n$ layers. 
``U-NET'' is utilized by adding skip connections between each layer $i$ and layer $n-i$, simply concatenating all channels of them respectively.
$n$ denotes the total number of layers of the generator.
Then the number of channels in decoder is changed, and the final decoder architecture can be expressed as:
$\mathit{TD_{512}}$-$\mathit{TD_{1024}}$-$\mathit{TD_{1024}}$-$\mathit{T_{1024}}$-$\mathit{T_{1024}}$-$\mathit{T_{512}}$-$\mathit{T_{256}}$-$\mathit{T_{128}}$.
The last layer in the decoder is followed by a convolution layer and a Tanh activation function. 
Finally, the generator produces the output maps.

As the generator is trained to produce out maps that cannot be distinguished from ground truth maps, 
we define the generator loss as a sigmoid cross-entropy loss of the generated images and an array of ones. 
The ground truth maps represent maps with $50$ feasible paths generated by the conventional RRT algorithm.
The function of sigmoid cross-entropy is defined as:
\begin{equation}\label{key}
sce(m, n) = 
-[n*\ln(M) + (1-n)*\ln(M)]
\end{equation}
\begin{equation}\label{key}
M = sigmoid(m) = \frac{1}{1 + \exp(-m)}
\end{equation}

The L1 loss is added for reducing blurring. Then the loss function of the generator is shown as follows:
\begin{equation}\label{key}
\mathit{L_{G}} = sce(G(o),1)  +\lambda\mathcal{L}_{L1}(G),
\end{equation}
where $\lambda$ is a weight coefficient. We set $\lambda=100$ in this article.
\subsection{Discriminator Architectures}
The discriminator architecture includes two pairs of input. 
One pair consists of the input maps and the corresponding ground truth maps, which should be classified as real.
Another pair consists of the input maps and the corresponding output maps, which should be classified as fake.
We concatenate each pair and perform encoding procedure respectively, using modules of the form convolution-Batchnorm-ReLU.
Then the architecture can be expressed as:
$\mathit{C_{64}}$-$\mathit{C_{128}}$-$\mathit{C_{256}}$-$\mathit{C_{512}}$-$\mathit{C_{512}}$-$\mathit{C_{512}}$-$\mathit{C_{512}}$-$\mathit{C_{512}}$.
All ReLU activation layers in the architecture are leaky, with slope $\alpha=0.2$.
The last layer is a one-dimensional output, followed by a Sigmoid function.

As the discriminator is trained to distinguish output maps and ground truth maps,
we define the discriminator loss as the sum of the real loss and the generated loss. 
The real loss is a sigmoid cross-entropy loss of the ground truth maps and an array of ones and generated loss is a sigmoid cross-entropy loss of the output maps and an array of zeros, respectively. 
Then the loss function of the discriminator is shown as follows:
\begin{equation}\label{key}
\mathit{L_{D}} = sce(G(o),0)  + sce(y,1),
\end{equation}
The generation procedure of ground truth images $y$ is described in Section V.

\subsection{CGAN-RRT*}
\begin{algorithm}[h]
	\normalsize
	\label{CGANRRT}
	\caption{The CGAN-RRT* Algorithm}
	$V$ $\gets$ \{$x_{init}$\}; $E$ $\gets$ $\emptyset$;
	\\
	$\mathcal{P} \gets \mathrm{CGANModel}(Map,y)$;
	\\
	\For {$i$=1,. . .,$n$}{
		\eIf{$Rand() \textless 0.5$}
		{$x_{rand} \gets \text{Nonuniform($\mathcal{P}$)}$}
		{$x_{rand} \gets \text{Uniform}_{i}$}
		{$x_{nearest}$} $\gets$ \text{Nearest}($G=(V,E)$,{$x_{rand}$});
		\\
		{$x_{new}$} $\gets$ \text{Steer}({$x_{nearest}$},{$x_{rand}$});
		\\
		\If{$\text{ObstacleFree}({x_{nearest}},{x_{new}})$}
		{$V$ $\gets$ $V$ $\cup$ \{{$x_{new}$}\}; 
		\\
		$E$ $\gets$ $E$ $\cup$ \{{($x_{nearest}$,$x_{new}$)}\};
		\\
		Rewire();
		\\
		
		}
	}
	\textbf{Return} ($G=(V,E)$)
\end{algorithm}
The CGAN-RRT* algorithm just differs from the traditional RRT* algorithm in the sampling technique.
Thus, it can be applied to other sampling-based algorithms in future research.
It utilizes the pre-trained conditional-GAN model to guide the sampling process of the RRT* algorithm.
There are two sampling strategies in the CGAN-RRT* algorithm.
One is the nonuniform sampling strategy guided by the predicted possibility distribution of feasible paths generated from the conditional-GAN model,
and another is the uniform sampling strategy adopted by the conventional RRT* algorithm.

The details of the CGAN-RRT* algorithm is presented in Algorithm $2$. 
It is initialized with the initial state $x_{init}$ and no edges.
$\mathrm{CGANModel}(Map,y)$ produces the predicted sampling distribution of feasible paths $\mathcal{P}$.
At each iteration, a random number $Rand() \in (0,1)$ is utilized to determine which sampling strategies to choose.
If $Rand() \textless 0.5$, a point $x_{rand}$ is sampled from $\text{Nonuniform($\mathcal{P}$)}$.
Otherwise, the uniform sampling strategies is utilized.
Then the algorithm attempts to find the nearest vertex $v \in V$ to the new sample and connect them.
If there are no obstacles through the connection, $x_{new}$ is added to the vertex set $V$,
and ($x_{nearest},x_{new}$) is added to the edge set $E$.
The procedure of $Rewire$ enables the initial path to converge to the optimal path.
Finally, the CGAN-RRT* algorithm returns a graph of a new vertex set $V$ and an edge set $E$.

\section{Simulation Results}
\label{sec:Simulation Results}

In this section, we first introduce the training details of our CGAN model in section \ref{sec:Simulation Results}-A.
Then we present the possibility distribution of feasible paths generated from the CGAN model and evaluate its performance in section \ref{sec:Simulation Results}-B.
Finally, we compare the CGAN-RRT* algorithm with the conventional RRT* algorithm through several experimental simulations in sections \ref{sec:Simulation Results}-C.

\subsection{Training Details}
Inspired by other GAN papers, we train the generator to maximize $logD(o,G(o,z))$, instead of minimizing $log(1-D(o,G(o,z)))$.
We train our CGAN model on the google co-laboratory platform with NVIDIA TESLA T4 and TensorFlow framework.
The Adam solver\cite{da2014method} and mini-batch stochastic gradient descent(SGD) are applied in the training process with a learning rate of $\alpha=0.0002$.
In addition, we utilize the suggested momentum parameters $\beta_{1} = 0.5$ and $\beta_{2} = 0.999$.

In order to guarantee the divergence and robustness of our CGAN model, $12,000$ $2$-D maps with randomly start points and endpoints are generated.
The training set includes three types of maps.
The size of each map is $(256, 256, 3)$.
Then we utilize the RRT algorithm to generate $50$ feasible paths on each map.
Finally, we combine the input maps and corresponding ground truth maps together as our training set.
The input maps are the original maps with randomly start and goal points, 
while the ground truth maps are maps with $50$ feasible paths generated from the RRT algorithm.
Our CGAN model is trained and tested with Python $3.7$.

\subsection{Evaluation of the CGAN Model}

\begin{figure}[h]
	\captionsetup{font={small}}
	\centering
	\includegraphics[width=1\linewidth]{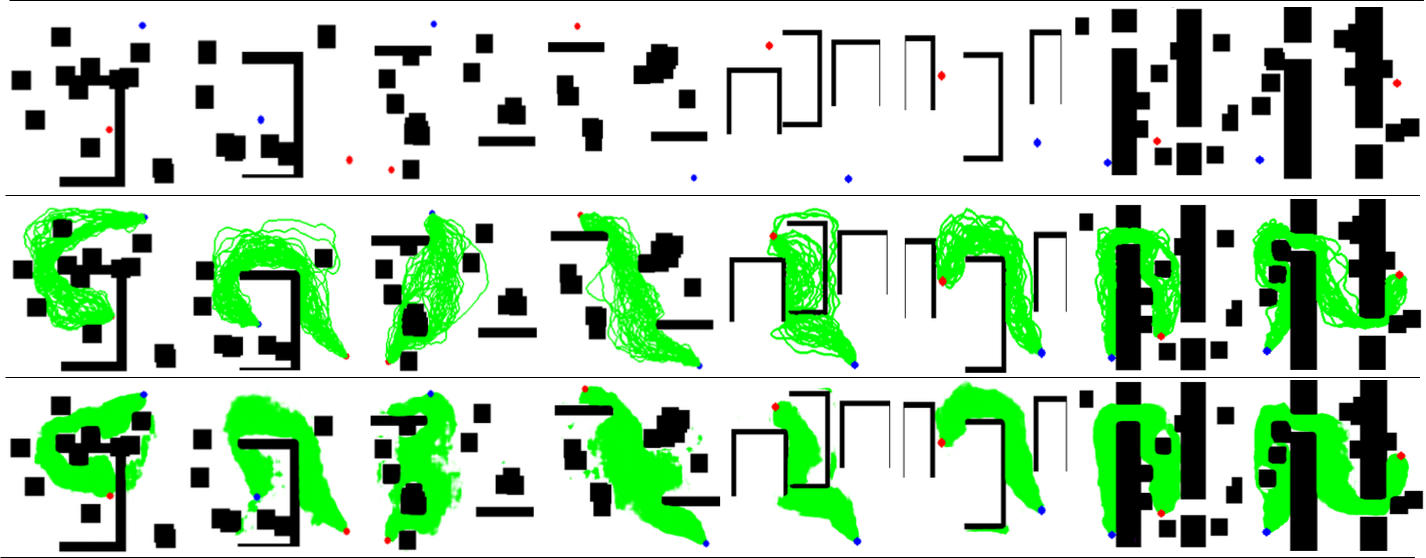}
	\caption{Examples of the proposed CGAN model.
			 The first row represents four types of input maps with start point and goal point.
			 The second row represents the corresponding ground truth maps with 50 feasible paths generated from the RRT algorithm.
			 The third row represents the corresponding output maps with predicted possibility distribution of feasible paths generated from the CAGN model.
			}
	\label{fig4}
\end{figure}

We generate two groups of combined raw maps and maps with generated feasible paths as the test set.
Each group includes $500$ pairs of images with a size of $(256, 256, 3)$.
One test group consists of three types of maps which are same as the training set,
and another test group consists of two types of maps which are different from the training set.
It only takes the proposed CGAN model $1.5s$ to output a predicted possibility distribution of feasible paths for a given map.
Some examples of generated results from the CGAN model are shown in Fig. \ref{fig4}, which demonstrates the performance of the proposed CGAN-model.
The first and second row of Fig. \ref{fig4} represents the initial maps and maps with feasible paths generated from the RRT algorithm, respectively.
And the output maps with predicted possibility distribution of paths are presented in the third row.
The red and blue circles denote the start and goal points, respectively.
The green region of each map in the second row represents $50$ feasible paths generated from the RRT algorithm, which is denoted as the ground truth.
While in the third row, the green region of each map represents the predicted possibility distribution of feasible paths generated from the CGAN model.
Fig. \ref{fig4} illustrates that the predicted possibility distribution is close to the ground truth.

\begin{figure}[h]
	\captionsetup{font={small}}
	\setlength{\abovecaptionskip}{0.cm}
	\begin{subfigure}[b]{.3\linewidth}
		\includegraphics[width=1\linewidth]{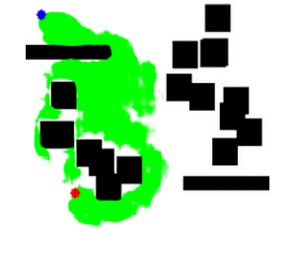}
		\caption{ }
	\end{subfigure}
	\hfill
	\begin{subfigure}[b]{.3\linewidth}
		\includegraphics[width=1\linewidth]{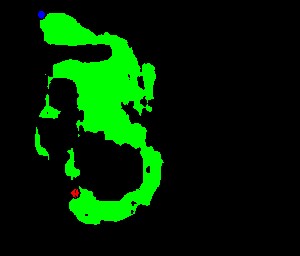}
		\caption{ }
	\end{subfigure}
	\hfill
	\begin{subfigure}[b]{.3\linewidth}
		\includegraphics[width=1\linewidth]{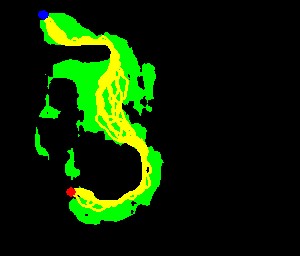}
		\caption{ }
	\end{subfigure}
	\hfill
	\caption{One example of the test process for generated possibility distribution of paths. (a): The Map with predicted possibility distribution of paths. (b): The map after color filter operation. Space except for paths, start point and goal point is turned to black color. (c): The map with paths generated from the RRT algorithm in the limited green space.}
	\label{fig5}
\end{figure}

\begin{figure*}[h]
	\centering
	\captionsetup{font={small}}
	\setlength{\abovecaptionskip}{0.cm}
	\begin{subfigure}[b]{.2\linewidth}
		\includegraphics[width=1.\linewidth]{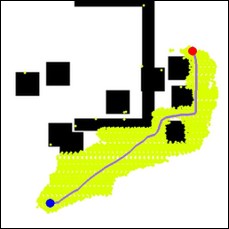}
		\caption{ }
	\end{subfigure}
	\hfill
	\begin{subfigure}[b]{.2\linewidth}
		\includegraphics[width=1.\linewidth]{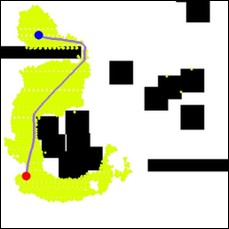}
		\caption{ }
	\end{subfigure}
	\hfill
	\begin{subfigure}[b]{.2\linewidth}
		\includegraphics[width=1.\linewidth]{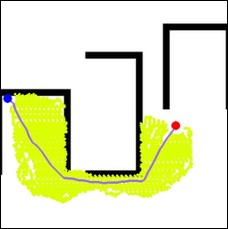}
		\caption{ }
	\end{subfigure}
	\hfill
	\begin{subfigure}[b]{.2\linewidth}
		\includegraphics[width=1.\linewidth]{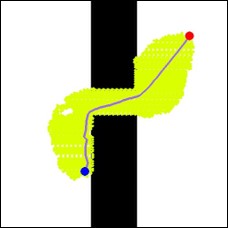}
		\caption{ }
	\end{subfigure}
	\caption{Optimal paths generated with predicted possibility distribution for four types of maps.
		The red and blue circles denote the start point and goal point, respectively. 
		The green-yellow represents possibility distribution from the CGAN model and medium-orchid represents the optimal path.
		(a): Map1. (b): Map2. (c): Map3. (d): Map4.}
	\label{fig6}
\end{figure*}

A variant of the RRT algorithm is utilized to test the connectivity of paths generated from the CGAN-model.
We set the region of predicted possibility distribution of paths as the free space $\mathcal{X}_{free}$.
Other regions are set as collision space.
All sampling process is operated in the free space.
If the RRT algorithm can find feasible paths connecting the start point and goal point in this space,
the predicted possibility distribution is viewed as feasible and effective.
One example of this process is shown in Fig. \ref{fig5}.
The red and blue circles denote the start and goal points, respectively.
We use green to represent the free space and black to represent the collision space.
Feasible paths generated from the RRT algorithm is represented by yellow in Fig. \ref{fig5}.
The test group consists of three types of maps which are same as the training set achieves $91.8\%$ success rate,
while another test group consisting of two types of maps which are different from the training set achieves $77.8\%$ success rate.
The test results illustrate that the proposed CGAN model is effective for generating the possibility distribution of feasible paths.
And the performance of the test group which consists of two types of maps different from the training set can be improved in future research.

\subsection{Comparison between CGAN-RRT* and RRT*}

In this section, we test the performance of CGAN-RRT* and RRT* algorithm on the path planning problem.
The CGAN-RRT* algorithm utilizes the predicted probability distribution of feasible paths to guide the sampling process of the RRT* algorithm,
while the conventional RRT* algorithm just uniformly samples points from the free space $\mathcal{X}_{free}$.
The step size of the two algorithms is both set to $6$.

The illustrations of the predicted possibility distribution and optimal path for four maps with different types are shown in Fig. \ref{fig6}.
The red and blue circles denote the start and goal points, respectively.
The yellow represents the predicted possibility distribution of paths generated by the conditional-GAN model, 
which indicates that the corresponding region has a higher possibility to contain the optimal path than other regions.
And we utilize medium-orchid color to represent the optimal path.
From the illustrations in Fig. \ref{fig6}, we can find that the optimal path lies in the predicted possibility distribution,
which indicates that the prediction results generated from the proposed conditional-GAN model can accelerate the convergence to the optimal path and improve the quality of the initial path.

These four maps with different types are utilized to test the performance of CGAN-RRT* and RRT* algorithm.
The time cost, the number of nodes, and the length of the initial path are selected as metrics to evaluate the performance.
The time cost refers to the execution time of the whole algorithm.
For the CAGN-RRT* algorithm, the time cost includes two parts. 
One part is the execution time of generating predicted probability distribution from the CAGN model, 
and another part is the execution time of the CGAN-RRT* algorithm with nonuniform sampling strategy guided by the predicted possibility distribution and uniform sampling strategy adopted by the conventional RRT* algorithm.
of which the nonuniform sampling strategy is guided by the predicted probability distribution and uniform sampling strategy is adopted 
The number of nodes when the optimal path is found is an essential evaluation metric since it represents the memory usage of path planning algorithms.
The quality of the initial path has a huge impact on path planning algorithms.
If the length of the initial path is smaller, the performance of algorithms tend to be more efficient.

By utilizing four maps in Fig. \ref{fig6}, we provide a statistical result in TABLE \ref{table} to demonstrate the comparison of performance between the proposed CGAN-RRT* algorithm and conventional RRT* algorithm 
It illustrates that the CGAN-RRT* algorithm achieves better performance compared with the RRT* algorithm.
First, the CGAN-RRT* algorithm consumes much less time to converge to the optimal path.
It means that the CGAN-RRT* algorithm can improve the operational efficiency of the path planning problem.
Second, a much smaller number of nodes are utilized by the CGAN-RRT* algorithm compared with the RRT* algorithm, 
which reveals that the CGAN-RRT* algorithm utilizes much less memory usage and can save a lot of computation resources.
Third, the initial path generated from the CGAN-RRT* algorithm is shorter than the initial path generated from the RRT* algorithm.
It means that the initial path generated from the CGAN-RRT* algorithm is closer to the optimal path.
TABLE I shows that the CGAN-RRT* algorithm can always find a high-quality initial path.

\begin{table}[htbp]
	\centering
	\caption{Comparison of Algorithm Performance}
	\renewcommand{\arraystretch}{1.3}
	\setlength{\tabcolsep}{1mm}{
		\begin{tabular}{c|c|c|c|c|c}
			\hline
			\multicolumn{1}{c}{} &       
			& \multicolumn{1}{p{5em}<{\centering}|}{{Time cost(s)}} 
			& \multicolumn{1}{p{5em}<{\centering}|}{{Number of nodes}} 
			& \multicolumn{1}{p{4.78em}<{\centering}|}{{Length of Ini. path}} 
			& \multicolumn{1}{p{5em}<{\centering}}{{Length of Opt. path}} \\
			\hline
			\multirow{2}[2]{*}{\text{Map1}} & \text{CGAN-RRT*} & $\textbf{136}$ & $\textbf{7272}$ & $\textbf{552}$ & \multirow{2}[2]{*}{$527$} \\
			& \text{RRT*} & $708$   & $17924$ & $589$   &  \\
			\hline
			\multirow{2}[2]{*}{\text{Map2}} & \text{CGAN-RRT*} & $\textbf{192}$ & $\textbf{9312}$ & $\textbf{504}$ & \multirow{2}[2]{*}{$435$} \\
			& \text{RRT*} & $419$   & $13680$ & $815$   &  \\
			\hline
			\multirow{2}[2]{*}{\text{Map3}} & \text{CGAN-RRT*} & $\textbf{206}$ & $\textbf{8876}$ & $\textbf{573}$ & \multirow{2}[2]{*}{$535$} \\
			& \text{RRT*} & $1229$  & $23006$ & $608$   &  \\
			\hline
			\multirow{2}[2]{*}{\text{Map4}} & \text{CGAN-RRT*} & $\textbf{75}$ & $\textbf{5170}$ & $\textbf{444}$ & \multirow{2}[2]{*}{$420$} \\
			& \text{RRT*} & $710$   & $18164$ & $513$   &  \\
			\hline
	\end{tabular}}
	\label{table}%
\end{table}%

In conclusion, the proposed conditional-GAN model can generate the predicted possibility distribution of feasible paths in a short time.
The success rate of two different test groups reveals that the model is effective.
By utilizing the nonuniform sampling strategy guided by the predicted possibility distribution generated from the proposed CGAN model,
the CGAN-RRT* algorithm achieves better performance compared with the RRT* algorithm in terms of the time cost, the number of nodes, and the length of the initial path.

\section{Conclusions and Future Work}
\label{sec:conclusion}

In this article, we present a novel algorithm CGAN-RRT* for the robotic path planning problem. 
It utilizes the predicted possibility distribution of paths produced from the CGAN model to guide the sampling process of the RRT* algorithm.
The CGAN model learns a number of maps with feasible paths generated from the RRT algorithm.
The simulation results in Section V suggest that the CGAN-RRT* algorithm is a promising approach for the path planning problem 
and it has a much better performance compared with the conventional RRT* algorithm in terms of the time cost, the number of nodes, and the initial path length.
Therefore, the proposed CGAN-RRT* algorithm can accelerate the convergence to the optimal path and provide a high-quality initial path.

As the CGAN-RRT* algorithm only differs from the conventional RRT* algorithm in terms of sampling strategies, it can be applied to other sampling-based path planning algorithms to improve the performance of them. 
Furthermore, the CGAN-RRT* algorithm can be extended and used in high-dimensional configuration space for path planning problems in future research.








\section*{Acknowledgment}
This project is supported by Shenzhen Science and Technology
Innovation projects JCYJ20170413161616163, Hong
Kong ITC ITSP Tier 2 grant \# ITS/105/18FP, Hong Kong
ITC MRP grant \# MRP/011/18 and Hong Kong RGC GRF
grant \# 14200618 awarded to Max Q.-H. Meng.

\ifCLASSOPTIONcaptionsoff
  \newpage
\fi




\bibliographystyle{IEEEtran}
\bibliography{IEEEabrv,refs}

%

\begin{IEEEbiography}
	[{\includegraphics[width=1in,height=1.25in,clip,keepaspectratio]{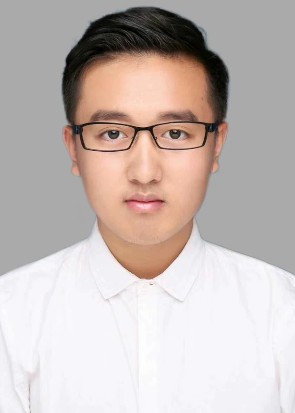}}]{Nachuan Ma}
	received the B.E degree in Electrical Engineering and Automation from China University of Mining and Technology, Xuzhou, China, in 2019,
	and the M.Sc. degree in Electronic Engineering from The Chinese University of Hong Kong, Hong Kong, in 2020.
	He is currently a Research Assistant with the Department of Electronic and Electrical Engineering of the Southern University of Science and Technology, Shenzhen, China.
	His current research interests include motion planning and simultaneous localization and mapping (SLAM).
\end{IEEEbiography}

\begin{IEEEbiography}
	[{\includegraphics[width=1in,height=1.25in,clip,keepaspectratio]{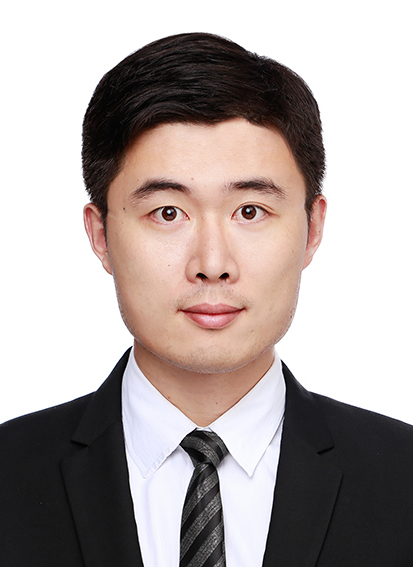}}]{Jiankun Wang}
	received the B.E. degree in Automation from Shandong University, Jinan, China, in 2015, and the Ph.D. degree in Department of
	Electronic Engineering, The Chinese University of Hong Kong, Hong Kong, in 2019. He is currently a Research Assistant Professor with the Department of Electronic and Electrical Engineering of the Southern University of Science and Technology, Shenzhen, China.
	
	During his Ph.D. degree, he spent six months at Stanford University, CA, USA, as a Visiting Student Scholar supervised by Prof. Oussama Khatib. His current research interests include motion planning and control, human robot interaction, and machine learning in robotics.
\end{IEEEbiography}

\begin{IEEEbiography}
	[{\includegraphics[width=1in,height=1.25in,clip,keepaspectratio]{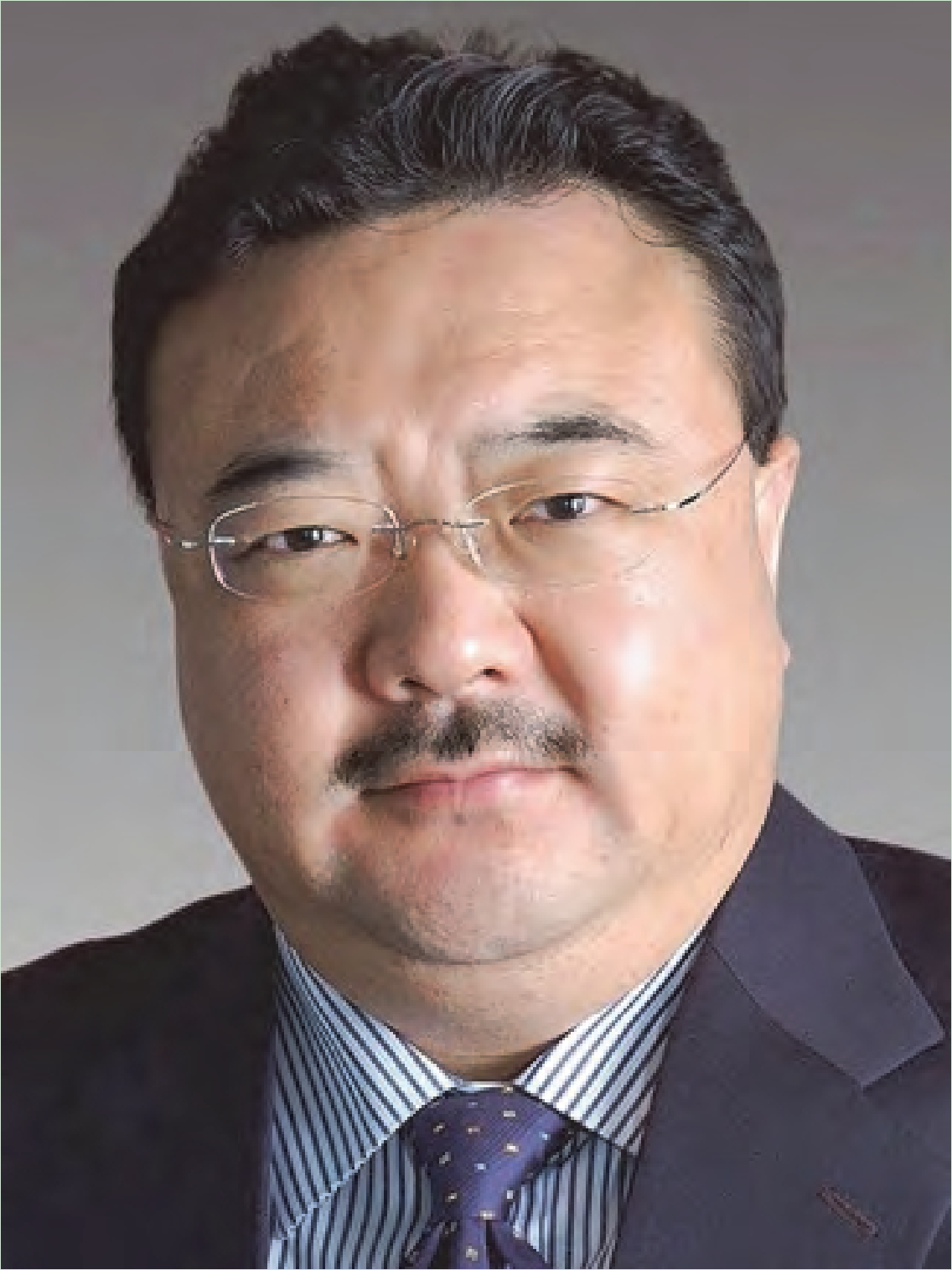}}]{Max Q.-H. Meng}
	received the Ph.D. degree in electrical and computer engineering from the University of Victoria, Victoria, Canada, in 1992.
	
	He is a chair professor with the Department of Electronic and Electrical Engineering of the Southern University of Science and Technology in Shenzhen, China, on leave from the Department of Electronic Engineering, The Chinese University of Hong Kong, Hong Kong, and also with the Shenzhen Research Institute of the Chinese University of Hong Kong in Shenzhen, China.
	He holds honorary positions as a Distinguished Professor with State Key Laboratory of Robotics and Systems, Harbin Institute of Technology, Harbin,
	China; a distinguished Provincial Professor with Henan University of Science and Technology, Luoyang, China; and the Honorary Dean of the School of Control Science and Engineering, Shandong University, Jinan, China. 
	His research interests include robotics, perception and sensing, human–robot interaction, active medical devices, biosensors and sensor networks, and adaptive and intelligent systems. He has published more than 500 journal and conference papers and served on many editorial boards.
	
	Dr. Meng is serving as an Elected Member of the Administrative Committee of the IEEE Robotics and Automation Society. He received the IEEE Third Millennium Medal Award.
\end{IEEEbiography}








\end{document}